\documentclass[sigconf]{aamas}

\usepackage{balance} 
\usepackage{subcaption}
\usepackage{dsfont}
\usepackage{pifont}
\newcommand{\cmark}{\textcolor{teal}{\ding{51}}}%
\newcommand{\xmark}{\textcolor{red}{\ding{55}}}%
\newcommand{\halfcmark}{\textcolor{orange}{\ding{51}}}
\newcommand{\nmark}{\textcolor{black}{\textbf{--}}}%

\makeatletter
\gdef\@copyrightpermission{
  \begin{minipage}{0.2\columnwidth}
   \href{https://creativecommons.org/licenses/by/4.0/}{\includegraphics[width=0.90\textwidth]{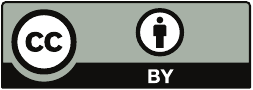}}
  \end{minipage}\hfill
  \begin{minipage}{0.8\columnwidth}
   \href{https://creativecommons.org/licenses/by/4.0/}{This work is licensed under a Creative Commons Attribution International 4.0 License.}
  \end{minipage}
  \vspace{5pt}
}
\makeatother

\setcopyright{ifaamas}
\acmConference[AAMAS '25]{Proc.\@ of the 24th International Conference
on Autonomous Agents and Multiagent Systems (AAMAS 2025)}{May 19 -- 23, 2025}
{Detroit, Michigan, USA}{Y.~Vorobeychik, S.~Das, A.~Nowé  (eds.)}
\copyrightyear{2025}
\acmYear{2025}
\acmDOI{}
\acmPrice{}
\acmISBN{}

\acmSubmissionID{1139}

\title[]{Discovery and Deployment of Emergent Robot Swarm Behaviors via 
Representation Learning and Real2Sim2Real Transfer}

\author{Connor Mattson}
\affiliation{
  \institution{University of Utah}
  \city{Salt Lake City, UT}
  \country{USA}}
\email{c.mattson@utah.edu}

\author{Varun Raveendra}
\affiliation{
  \institution{University of Utah}
  \city{Salt Lake City, UT}
  \country{USA}}
\email{varun.raveendra@utah.edu}

\author{Ricardo Vega}
\affiliation{
  \institution{George Mason University}
  \city{Fairfax, VA}
  \country{USA}}
\email{rvega7@gmu.edu}

\author{Cameron Nowzari}
\affiliation{
  \institution{George Mason University}
  \city{Fairfax, VA}
  \country{USA}}
\email{cnowzari@gmu.edu}

\author{Daniel S. Drew}
\affiliation{
  \institution{University of Hawaii at Manoa}
  \city{Honolulu, HI}
  \country{USA}}
\email{ddrew@hawaii.edu}

\author{Daniel S. Brown}
\affiliation{
  \institution{University of Utah}
  \city{Salt Lake City, UT}
  \country{USA}}
\email{daniel.s.brown@utah.edu}

\begin{abstract}
Given a swarm of limited-capability robots, we seek to automatically discover the set of possible emergent behaviors. Prior approaches to behavior discovery rely on human feedback or hand-crafted behavior metrics to represent and evolve behaviors and only discover behaviors in simulation, without testing or considering the deployment of these new behaviors on real robot swarms. In this work, we present Real2Sim2Real Behavior Discovery via Self-Supervised Representation Learning, which combines representation learning and novelty search to discover possible emergent behaviors automatically in simulation and enable direct controller transfer to real robots. First, we evaluate our method in simulation and show that our proposed self-supervised representation learning approach outperforms previous hand-crafted metrics by more accurately representing the space of possible emergent behaviors. Then, we address the reality gap by incorporating recent work in sim2real transfer for swarms into our lightweight simulator design, enabling direct robot deployment of all behaviors discovered in simulation on an open-source and low-cost robot platform. 
\end{abstract}

\keywords{Swarm Robotics; Emergent Behaviors; Representation Learning}

\newcommand{\BibTeX}{\rm B\kern-.05em{\sc i\kern-.025em b}\kern-.08em\TeX}

\begin{document}


\pagestyle{fancy}
\fancyhead{}


\maketitle 


\section{Introduction}

\begin{figure}
    \centering
    \includegraphics[width=0.95\linewidth]{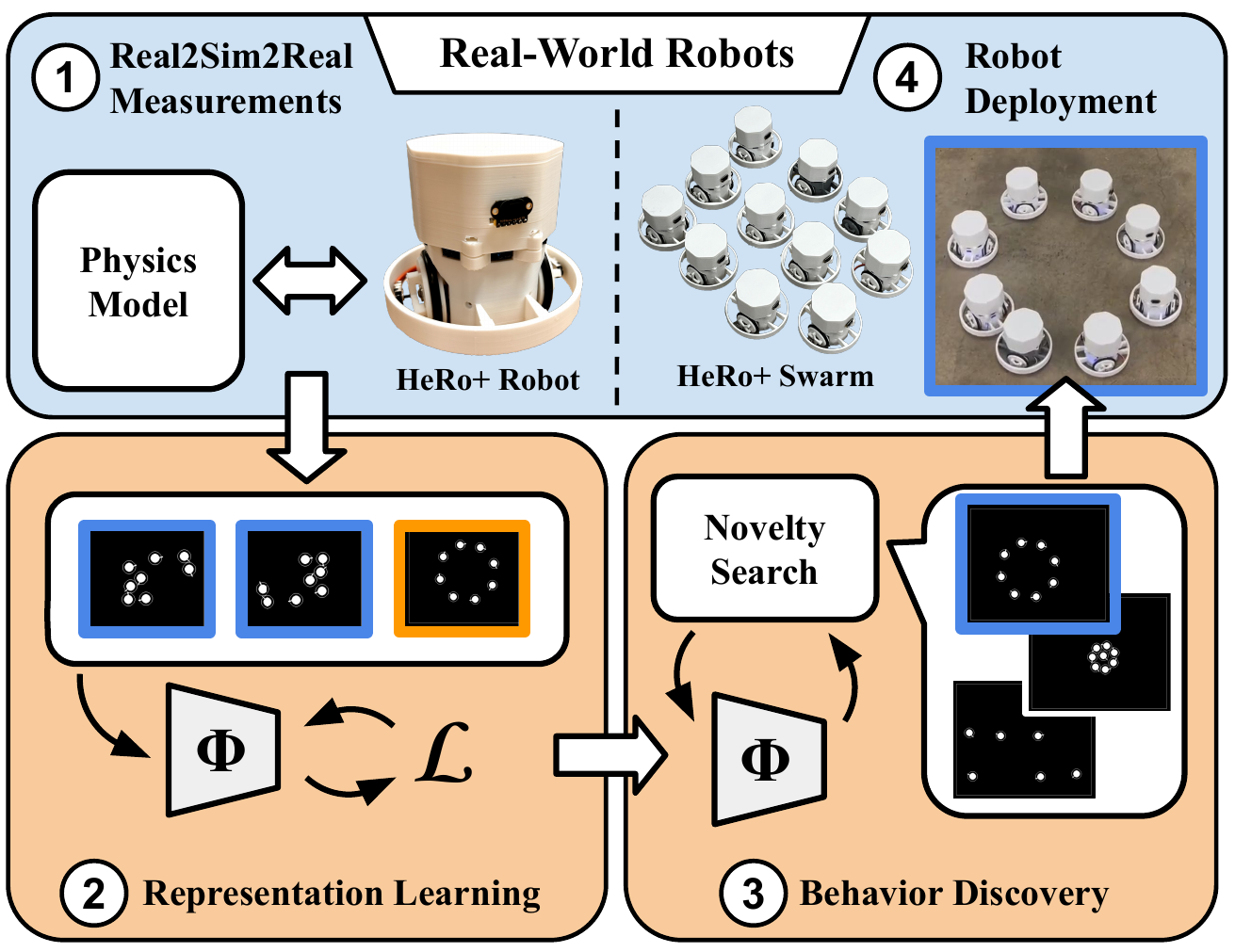}
    \caption{Real2Sim2Real Behavior Discovery via Self-Supervised
Representation Learning
    \normalfont discovers new collective behaviors for robot swarms while addressing the Sim2Real gap. \textbf{1)} Real robot  measurements are implemented into a physics model in software. \textbf{2)} The model is used to generate thousands of randomly sampled behavior videos, which are used to train a representation encoder using self-supervised learning. \textbf{3)} The trained encoder is used to discover novel emergent behaviors. \textbf{4)} Interesting behaviors found in simulation can be deployed on real robots without the need for controller adjustment. 
    }
    \label{fig:teaser}
\end{figure}

Decentralized robot swarms have the potential to provide efficient, low-cost, and robust solutions to tasks such as precision agriculture~\cite{blender_managing_2016, albani_field_2019}, aquatic monitoring~\cite{duarte_evolution_2016, berlinger_implicit_2021}, search and rescue~\cite{arnold_search_2018}, construction~\cite{petersen2011termes}, and object transportation~\cite{wilson_multi-robot_2018, gauci2014clustering, alkilabi_cooperative_2017, chen_strategy_2013}. However, as robots become empowered with additional sensing, computation, and actuation capabilities, humans are faced with increasing cognitive complexity as they try to understand how their robots can work together to accomplish real-world goals. Researchers and practitioners currently lack the ability to fully understand and explore the design space of possible emergent swarm behaviors.  In this paper, we propose and evaluate a novel pipeline for automatically discovering the set of emergent behaviors that can be achieved for swarms of robots with a set of limited capabilities. Importantly, we leverage real2sim2real techniques~\cite{lim2021planar,vega_simulate_2023} so that the behaviors we discover can be directly deployed in the real world.




Most prior research on swarm robotics focuses on optimizing swarm behaviors for specific tasks. Extensive work has been performed to optimize swarm controllers to produce target behaviors such as aggregation~\cite{arvin_imitation_2011, garnier_embodiment_2008, kernbach_adaptive_2013, timmis_immune-inspired_2016, gauci_self-organized_2014, feola_aggregation_2023, gauci2014evolving}, circle formation~\cite{jung2013shaping,st2018circle, zhu2024spiking}, chain formation~\cite{rezeck_chemistry-inspired_2022, sperati2011self}, milling~\cite{brown2014human,berlinger_implicit_2021}, spatial coverage~\cite{ozdemir_spatial_2019}, self-assembly~\cite{gross_autonomous_2006}, segregation~\cite{rezeck_flocking-segregative_2021, santos_2020_spatial} and shepherding~\cite{ozdemir2017shepherding}, many of which have been successfully deployed on real robots. By contrast, rather than seeking a specific pre-imagined behavior, we explore the more open-ended problem of discovering the set of emergent behaviors that are possible given a particular robot swarm.

Recent work has started to address this problem by proposing approaches based on novelty search~\cite{lehman2011abandoning} for developing a taxonomy of possible emergent behaviors~\cite{brown_discovery_2018, mattson_exploring_2023, mattson_leveraging_2023}. Prior work, however, is limited to simulation and does not consider the challenges involved in transferring behaviors discovered in simulation to the real world. As we show in Section~\ref{sec:res-ablation}, ignoring the sim2real gap results in the discovery of behaviors that only reliably work in simulation. By contrast, we seek to address the challenging sim2real gap~\cite{jakobi1995noise, koos2012transferability} in order to discover emergent swarm behaviors that are actually deployable on real robots. Furthermore, prior work on swarm behavior discovery leverages large amounts of human effort in terms of carefully hand-constructed representations of swarm behaviors. Instead, we evaluate an entirely self-supervised approach to behavior discovery, where low-dimensional representations of high-dimensional behaviors are learned without the need for human fine-tuning or physics-based hand-crafted representations. 


We introduce \textbf{Real2Sim2Real Behavior Discovery via Self-Supervised
Representation Learning}, a combination of sim2real transfer, behavior representation learning, and novelty search that seeks to discover the set of possible emergent swarm behaviors without the need for expensive human feedback or hand-crafted representations. As shown in Figure~\ref{fig:teaser}, our method starts with the real-world robots and adopts the recently proposed \textit{Reality-to-Simulation-to-Reality for Swarms (RSRS)}  process~\cite{vega_simulate_2023} to systematically approximate and implement measured robot dynamics in a simulator, and then test that behaviors produced in our adapted simulation can be accurately reproduced in the real world.
Given an improved simulator, we then perform self-supervised representation learning to learn latent representations of videos of swarm behaviors. Finally, we use the learned representations to perform novelty search~\cite{lehman2011abandoning} to efficiently explore the space of emergent behaviors. By augmenting behavior discovery with the RSRS process, we enable direct deployment of all behaviors discovered in simulation to our open-source real-world HeRo+ Robots, an improved version of the educational HeRo~\cite{rezeck_hero_2023} robot. 

The contributions of this work can be summarized as follows: First, we propose a novel self-supervised representation learning approach for swarms based on SimCLR~\cite{chen2020simple} and demonstrate that it enables quantitatively better representation learning for swarm behaviors when compared to the hand-crafted behavior representations used in prior work~\cite{brown_discovery_2018}.

Second, we improve the open-source HeRo~\cite{rezeck_hero_2023} robot hardware to enable accurate time-of-flight sensing, be more robust to collisions, and reduce encoder error, resulting in improved hardware for swarm robotics and a reliable way to implement inexpensive line-of-sight sensing. We make our modified robot, HeRo+, open-source for other researchers to deploy.
    
Third, we demonstrate the first deployment and evaluation of emergent behavior discovery for robot swarms by augmenting behavior discovery with real2sim2real calibration~\cite{vega_simulate_2023}. 
Inspired by prior work on computation-free swarms~\cite{gauci2014clustering}, we define a simple line-of-sight capability model and show that our method automatically discovers deployable emergent swarm behaviors for aggregation, cyclic pursuit, and dispersal.
By contrast, behaviors discovered via unaligned simulations (where sensing and actuation parameters are not tuned based on real-world observations) have a much lower chance of working in the real world.
    

Finally, we highlight the practical considerations for deploying multi-robot systems while also paving the way for researchers to more easily discover and explore the space of emergent behaviors that are possible given a swarm of robots.




\section{Problem Statement} \label{sec:ProblemStatement}
Following existing nomenclature~\cite{brown_discovery_2018, mattson_exploring_2023}, we define our robots as agents with a well-defined capability model, $C = \left<S, M, A \right>$ composed of sensing ($S$), memory ($M$), and actuation ($A$) capabilities. In this paper, we seek to answer the following research question: \textbf{Given $\mathbf{N}$ robots with capabilities $\mathbf{C}$, what is the complete set of emergent behaviors that can be deployed on these robots?}

We model this problem as a search for a set of emergent behaviors in a behavior space $\mathcal{B}$. 
The difficulty of this problem stems from the assumption that we have no direct access to this behavior space. Instead, we seek to sample and simulate swarm controllers and infer their behavioral characteristics based on the visual output of a simulator. 
We assume that we know the space of possible swarm controllers, $U(C)$, and the swarm's environment, $\mathcal{E}$. The controller space and environment form the input parameters for a behavior map, $\phi : U(C) \ \times \ \mathcal{E} \to \mathcal{B}$, that returns a behavior representation in the space $\mathcal{B}$. While prior work has assumed access to a known function $\phi$~\cite{brown_discovery_2018}, we consider the case where there is no predefined knowledge of how to represent behavior characteristics in a low-dimensional space where search can be performed. 

\section{Methods}
The goal of our work is to leverage machine learning to learn low-dimensional latent representations of swarm behaviors, then use that model as the basis for exploration in search of new swarm behaviors that can be deployed on real robots. Our work differs from prior work in that it learns a latent representation model in an entirely self-supervised manner and our work leverages recent work in Swarm Real2Sim2Real transfer~\cite{vega_simulate_2023}, enabling direct deployment of discovered emergent behaviors to a real swarm of robots.

In the following subsections, we describe Real2Sim2Real Behavior Discovery via Self-Supervised
Representation Learning which employs in-simulation representation learning (\ref{sec:method-repr-learning}), behavior exploration and discovery via novelty search~(\ref{sec:method-behavior-discovery}), and simulator design that enables rapid and reliable real-world deployment (\ref{sec:method-rsrs}).

\subsection{Representation Learning} \label{sec:method-repr-learning}
In order to discover new behaviors, we need a way to  be able to characterize and represent different swarm behaviors. In prior work on behavior discovery, behavioral representations were explicitly hand-crafted as functions of the robots' Cartesian position and velocity~\cite{brown_discovery_2018}. However, recent advancements in representation learning enable training networks to represent high-dimensional data (images, video, etc.) as low-dimensional latent vectors that contain encoded information about the original data. Rather than manually crafting behavioral characteristics, we study to what extent we can leverage unsupervised representation learning to create meaningful embeddings from videos of swarm behaviors.

\subsubsection{Learning Paradigm} To achieve both self-supervised training and sufficient representation learning, we employ the popular Simple Framework for Contrastive Learning of Visual Representations (SimCLR)~\cite{chen2020simple}. SimCLR is based on \textit{contrastive learning}, where representations are learned by comparing and/or contrasting pairwise or triplet elements of the training data. For example, a network could learn that two elements of the data that are similar should be embedded in close proximity within the latent space (and vice versa for data that are different). In our case, SimCLR samples a swarm behavior video, $\mathbf{x}$, from our video dataset, uses two data transformations to alter the visual appearance, denoted $\mathbf{\widetilde{x}_i}$ and $\mathbf{\widetilde{x}_j}$, and then optimizes a network to embed $\mathbf{\widetilde{x}_i}$ and $\mathbf{\widetilde{x}_j}$ closer together in the latent space while considering all other elements of a batch as dissimilar from $\mathbf{\widetilde{x}_i}$ and $\mathbf{\widetilde{x}_j}$. This results in the following loss function for a positive pair of elements in a batch of size $N$,
\begin{equation}
    \label{eq:NXent}
    \mathcal{L}_{i,j} = -\text{log}\frac{\text{exp}(\text{sim}(z_i, z_j) / \tau)}{\sum_{k=1}^{2N} \mathds{1}_{[k \neq i]} \text{exp}(\text{sim}(z_i, z_k)/ \tau)},
\end{equation}
where $z_{\{i, j, k\}}$ is the latent embedding of $\widetilde{x}_{\{i, j, k\}}$, \textit{sim} is a function that measures the similarity between two vectors (e.g., cosine similarity, L2 distance), $\tau$ is a temperature parameter, and $\mathds{1}_{[k \neq i]}$ is a function that evaluates to 1 if and only if $k \neq i$ and evaluates to 0 otherwise. This loss objective is a cross-entropy formulation that simply seeks to maximize the likelihood that the $i$th and $j$th embeddings have the highest measure of similarity when compared to all other elements in the batch.

\subsubsection{Data Augmentation} 
Chen et al.~\cite{chen2020simple} conducted a thorough analysis of which augmentations produced the highest performing learned representation, and several other studies have thoroughly explored the benefits of pixel-based data augmentation in machine learning~\cite{krizhevsky2012imagenet, bachman2019learning, xie2020unsupervised}. 
In our paper, we implement one of the highest scoring combinations of transformations studied in prior work: random crop followed by random rotation~\cite{chen2020simple}.

\subsubsection{Encoder/Projection Architecture}\label{sec:methods-repr-arch} 
Following training, we use our learned SimCLR encoder, $\mathcal{\phi}$, as a means of obtaining low-dimensional behavior representations of each swarm video. These representations are then used to determine how similar two behaviors are. We highlight the importance of correctly defined notions of similarity in the following section, where we describe how we use our encoder to search for novel emergent behaviors.

\subsection{Behavior Discovery} \label{sec:method-behavior-discovery}
We seek to automatically discover new behaviors, which is both a non-stationary and non-trivial objective. In particular, methods that are apt for approximating stationary functions using gradient-based approaches are unlikely to converge to solutions that truly explore the space of all possible behaviors. Instead, we follow prior work in behavior discovery~\cite{brown_discovery_2018, mattson_leveraging_2023, mattson_exploring_2023} by implementing an evolutionary approach to exploration problems called \textit{Novelty Search}~\cite{lehman2011abandoning}. Novelty search serves as a fitness mechanism that rewards representations that are different from all previously observed representations. To facilitate this, novelty search aggregates representations to a dynamic buffer, denoted $B$. For any newly observed representation, $b \in \mathds{R}^d$ in $d$-dimensional latent space, the novelty of $b$ is defined as
\begin{equation}
    \label{eq:novelty}
    \text{Novelty}(b, B) = \frac{1}{k} \sum_{i=0}^k \text{dist}(b, B_i),
\end{equation}
where $k$ is the number of nearest neighbors in $B$ to consider and $B_i$ is the $i$th nearest neighbor of $b$ in $B$.

In several studies, novelty search has been used to aid optimization problems by incentivizing exploration in tandem with an optimization objective~\cite{gomes2013evolution, lehman2010efficiently, lehman2011evolving}. By contrast, our sole objective is exploration and novelty, which enables us to use it as the only component of a fitness function for evolutionary search. Evolutionary search is commonly utilized in robotics literature as a means of gradient-free optimization~\cite{doncieux2015evolutionary}. Succinctly, evolutionary search attempts to maximize a fitness function $f(g)$ by genetically evolving populations of genomes, $g$, where the highest scoring genomes are more likely to mutate and survive across multiple generations. In the context of our approach, our fitness function is the novelty function (Eq.~\ref{eq:novelty}) and the genome is a parameterized swarm controller that can be represented as a vector $g \in U(C)$. Let the function $\mathcal{S}(g)$ denote the simulation of controller $g$ which returns a behavior video, $\mathbf{x}$. Then, with the representation encoder, $\phi$, the evolutionary optimization can be written in the form
\begin{equation}
\label{eq:objective_func}
\max_{g \in U(C)} \text{Novelty}( \phi( \mathcal{S}( g )), B ).
\end{equation}

As previously mentioned, the objective shown above is non-stationary in the sense that a genome will always return a higher novelty score in an earlier generation than the same genome would in a later generation. This means that solutions that have high fitness early in the search will not be as novel in subsequent generations, requiring the algorithm to test new genomes to try to diversify the behavior space. After search, the novelty buffer ($B$) is passed through a k-Medoids clustering algorithm and the resulting behaviors are returned to the user.

\subsection{Real2Sim2Real Simulator Design (RSRS)} \label{sec:method-rsrs}
We augment the problem formation for behavior discovery used in prior work by explicitly targeting direct sim2real deployment for our discovered behaviors. One natural way to enable this is to simply use a high-fidelity robot simulator that can model the physics of the real world with sufficient precision. While this solution may be appropriate for methods that are directly optimizing for a specific behavior, our problem requires evaluating thousands of swarm controllers for simulations with hundreds of timesteps. We seek to instead use a simulator that is lightweight, as has been utilized in prior literature for swarm simulators that require costly search~\cite{mattson_leveraging_2023, snyder2023zespol}. Simultaneously, we do not want to neglect the dynamics of the real world, as over-simplification may produce behaviors that are infeasible for hardware deployment. Therefore, we require a strategic simulator design approach that enables both lightweight evaluation and closes the reality gap.

Reality-to-Simulation-to-Reality for Swarms (RSRS)~\cite{vega_simulate_2023} is a new simulator design paradigm that leads to more feasible and reliable real-world swarm deployments. The main idea behind RSRS is that the robots in simulation are \textit{less} capable than they actually are in the real world. At first, this may appear as a weakness, as our problem is defined with respect to the capabilities of real world robots. Although the dynamics of the real world are difficult to efficiently represent in simulation, we can approximate these real-world uncertainties by exaggerating their impact on our robots. For example, the collision dynamics between two robots cannot be perfectly modeled in a lightweight simulator, but real robot collisions can guide our approach to simulator design through informative observations. For example, for the observation that ``these robots cannot reliably slide past each other if they collide head-on,'' this perhaps indicates that the friction coefficient should be adjusted to ensure that the simulator does not attempt to find behaviors that exploit collisions.  We follow the four steps described in the RSRS process~\cite{vega_simulate_2023}: 
\textbf{1)} Measure the capabilities and dynamics of real-world robots, 
\textbf{2)} Implement the measurement data into the robot simulator,
\textbf{3)} Run experiments in simulation, modify robots and simulator as needed,
\textbf{4)} Perform experiments on real robots, modify robots and simulator as needed.

In particular, it should be noted that steps 3 and 4 involve iterative refinement to the simulator and robots in order to reproduce behaviors in the real world. RSRS lays out a simple \textit{if-else} approach for modifying the simulator and robots. If it is less expensive to upgrade the robots to improve reliability than to modify the simulator, upgrade the robots. Otherwise, make the simulator more realistic and take more measurements on real robots. RSRS has been previously shown to be effective when optimizing sim2real transfer for a specific desired behavior~\cite{vega_simulate_2023}. By contrast, our approach expands the use of this design paradigm to broadly enable Real2Sim2Real transfer for open-ended behavior discovery. We discuss the details of these steps and discuss the modifications we made to our robots and simulator in Sections~\ref{sec:exp-robot-model} and~\ref{sec:exp-rsrs}. 

\begin{figure}
    \centering
    \includegraphics[width=0.85\linewidth]{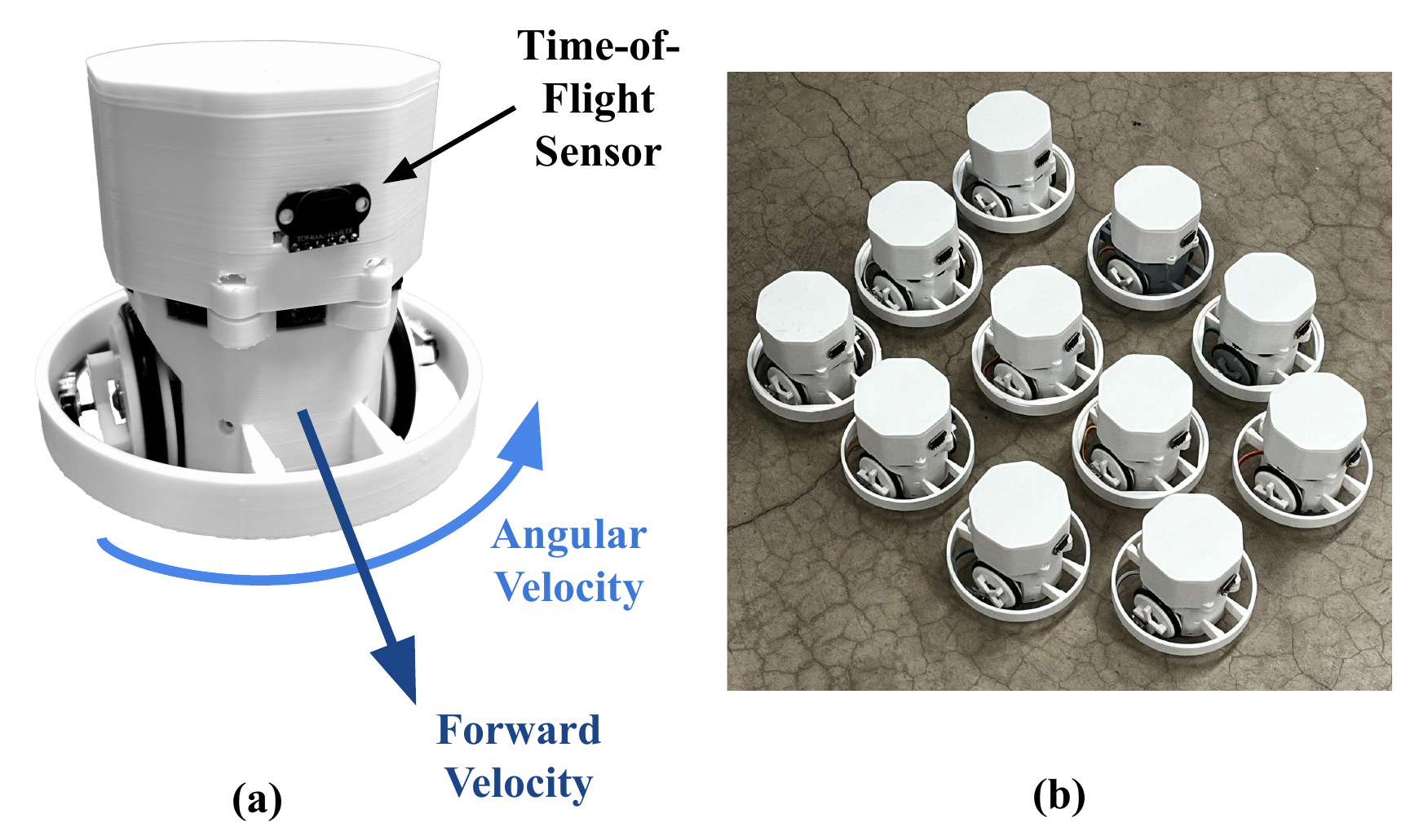}
    \caption{HeRo+ Robots: 
    \normalfont We deploy newly discovered behaviors on a fleet of HeRo+ robots. \textbf{(a)} A single HeRo+ robot uses unicycle commands to locomote and time-of-flight sensing to detect other robots. Our open-source robot design is mostly 3D-printed and costs approximately \$80-USD, making it an extremely low-cost option for swarms research.  \textbf{(b)} Our swarm of 11 HeRo+ Robots, 8 of which are used in this study.}
    \label{fig:herobots}
\end{figure}


\section{Experiments}
We demonstrate the efficacy of our methods by deploying new behaviors, discovered in simulation, on real swarm robots. Following our 4-stage RSRS process, we first model the interactions between our low-cost HeRo+ robots (Figure~\ref{fig:herobots}) in simulation using measurements obtained from the real world (\ref{sec:exp-robot-model}-\ref{sec:exp-rsrs}). Second, we generate a video dataset that is used to train a SimCLR encoder using self-supervised learning (\ref{sec:exp-repr-learning}), and employ evolutionary behavior discovery to evolve and discover unique behaviors that diversify the set of embedded features in the encoder (\ref{sec:exp-behavior-discovery}). Finally, the behaviors discovered in simulation are deployed on our robots in the real world (\ref{sec:exp-robot-deploy}). Videos, code, and open-source hardware designs are available on our project webpage
\footnote{
\url{https://sites.google.com/view/swarmdiscovery-with-rsrs/home}
}.

\subsection{Robot Hardware}\label{sec:exp-robot-model}


\subsubsection{Kinematics and Controllers}\label{sec:exp-robot-controllers}
Our experiments consider a homogeneous swarm of 8 robots modeled in 2D with unicycle kinematics, where the $i$th robot is controlled at time $t$ with a forward velocity, $v_{i, t}$, and angular velocity, $\omega_{i, t}$. Our robots only receive a binary observation $h_{i, t} \in \{0, 1\}$ from a line-of-sight sensor and use this signal as the condition for an \textit{if-else} style controller of the form $[u_{v, 0}, u_{\omega, 0}, u_{v, 1}, u_{\omega, 1}] \in U(C)$. Though the values of this controller are shared by all the agents in the swarm, the velocity of different agents may vary based on the agent's individual observation state,

{
\begin{align}
    (v_{i, t}, \omega_{i, t}) = \begin{cases}
                                     (u_{v, 0}, u_{\omega, 0}) & \text{if } h_{i, t} \ \text{is 0,} \\
                                     (u_{v, 1}, u_{\omega, 1}) & \text{otherwise.}
                                 \end{cases}
\end{align}
}
Because the 4-tuple controller representation is time-invariant, it allows for control of the behavior of the entire swarm, for an arbitrary simulation horizon, with just four scalar values. These four values, constrained by the robots' practical velocity limits, form the space of possible controllers, $U(C)$, where we search for behaviors as described in Section \ref{sec:method-behavior-discovery}. 


\subsubsection{Improvements to Robot Hardware}\label{sec:exp-robot-hardware-rsrs}
The HeRo Robot~\cite{rezeck_hero_2023} is an open-source, low-cost, 3D-printed robot with two-wheeled differential drive actuation. These robots act using only local observations and are effectively decentralized. However, for ease of deployment and control, the robots are connected to a centralized ROS server, allowing for full swarm emergency stop, synchronized start, and wireless controller updates.

Our first emergent behavior test on these robots was to attempt to get them to perform a ``Cyclic Pursuit'' behavior (all robots form a circle and rotate about the center) using already discovered unicycle controllers from prior work~\cite{vega_simulate_2023, mattson_leveraging_2023}. Recall from Section~\ref{sec:method-rsrs} that iteration for RSRS can take two forms: Hardware Upgrades and Software Upgrades. Using our observations from the real world, we close the reality gap from a hardware perspective by improving the HeRo hardware with the following features: 

\textbf{Bump Shield}: Based on our observations, the most significant hindrance to emergent behavior was that collisions between robots often resulted in actuation difficulties, especially when the contact was chassis-to-wheel (causing a direct force on the servo motors and wheels). For behaviors like cyclic pursuit, the robots may bump into each other during formation, which would effectively halt some robots in collision, resulting in a pile-up. Rather than attempting to carefully model the difference between head-on and chassis-to-wheel collisions, we found that augmenting robots with a 3D-printed bump shield was an inexpensive way to allow the robots to collide with each other without actuation faults. 

\textbf{Time-of-Flight Sensing}: By default, the HeRo robot uses 8 IR sensors, evenly spaced around the robot's circumference, to sense its surroundings. We tested our binary controller with just the forward-facing IR sensor and found that the sensor could not detect robots at a distance greater than 25~cm and reported false negatives 50\% of the time at the 25~cm range, which inhibited the ability of the robots to correctly sense each other and form together into a behavior. While these errors can be measured and modeled in simulation, we found that sensing reliability was critical to behavior formation, necessitating a hardware upgrade. We implement an inexpensive upgrade by adding a single laser-ranging Time-of-Flight sensor (VL53L1X, see Figure~\ref{fig:herobots}a) that can detect other robots up to 2~m away and is significantly more reliable, with almost no false negatives when the robots are driving at low speeds.

\textbf{Encoder Feedback}: Lastly, we found that the original position of the gear-driven encoders on the HeRo robot resulted in frequent reading errors that affected the reliability of low-level PID control. While it would be difficult to efficiently model this type of uncertainty in simulation, moving the encoder outside the wheels of the robot and connecting it with a directly-driven shaft significantly reduced the error frequency.

We found that these hardware improvements greatly increase the reliability of swarm controllers. We call this RSRS-improved HeRo robot, the \textbf{HeRo+ robot}. All CAD models and a bill of materials have been made available in our supplemental materials. 



\subsubsection{Environment:}
The robots are placed in a 170x142~cm arena. The four walls of the arena are each 5~cm tall and were deliberately designed so as to be shorter than the TOF sensor on the HeRo+ robots, preventing them from detecting the walls so that anything in line-of-sight can be assumed to be another agent. The environment also has 3 x 4 = 12 grid initialization points that are roughly centered in the arena. For simulated controllers, the 8 robots are randomly assigned to one of the 12 starting locations and randomly oriented in the range $[0, 2\pi]$ (following similar instantiation as other robot swarm studies~\cite{gauci_self-organized_2014}). When running tests on the real-robots, we replicate as closely as possible the same starting positions and orientations used in simulation.

\begin{figure*}
    \centering
    \includegraphics[width=0.82\linewidth]{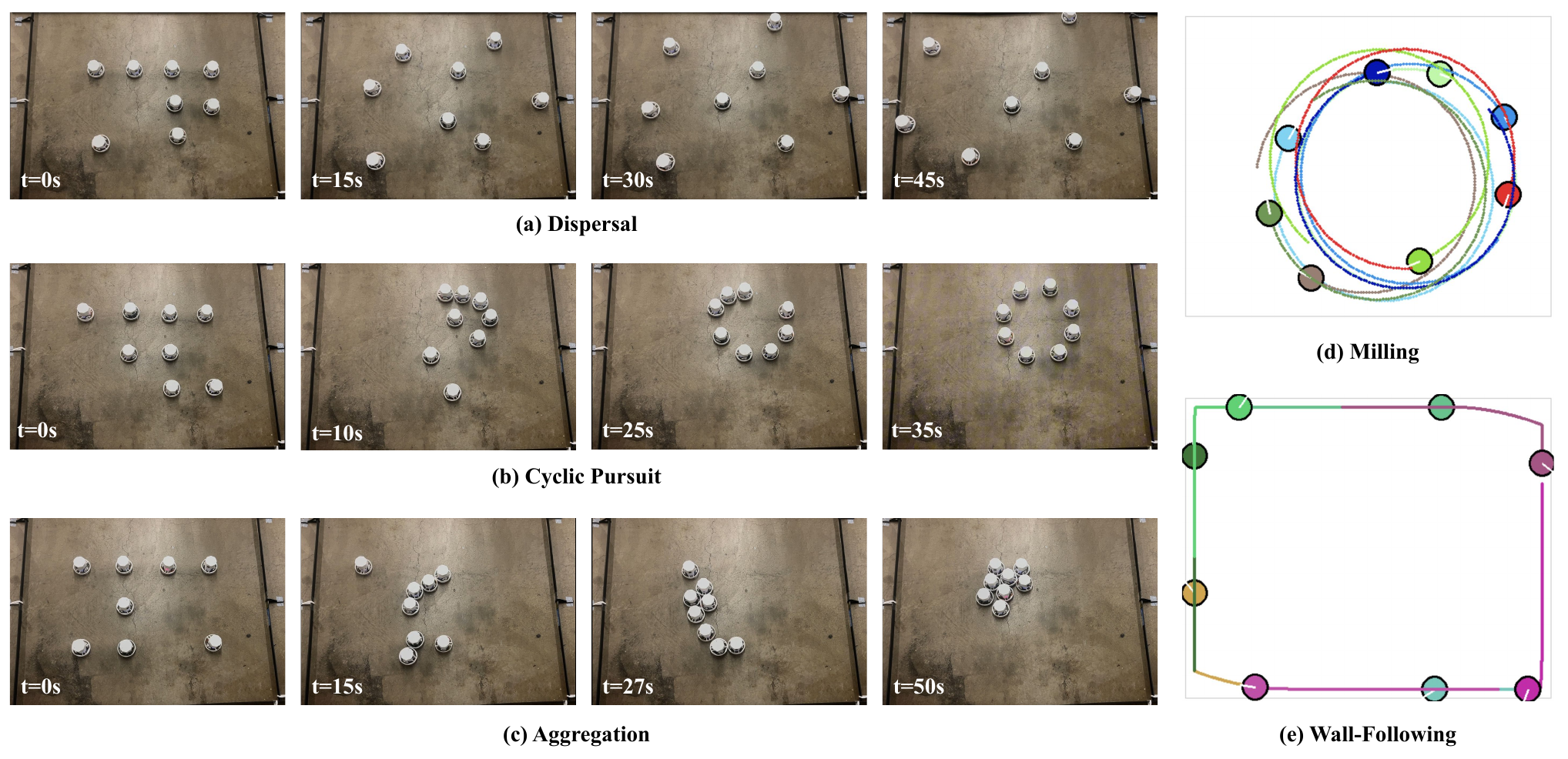}
    \caption{Discovered Behaviors.
    \normalfont Behaviors \textbf{a-c} were automatically discovered and deployed on HeRo+ Robots using Real2Sim2Real (RSRS) Representation Learning for Behavior Discovery. Behaviors \textbf{d-e} were discovered in simulation for a simulator without RSRS improvements, but could not be deployed to the real world and were not discovered when RSRS was added. \textbf{(a)} \textbf{Dispersal}: Agents cover the environment by moving away from other robots. \textbf{(b)} \textbf{Cyclic Pursuit}: Agents form a circular chain and revolve around the center. \textbf{(c)} \textbf{Aggregation}: Agents cluster together in the middle of the environment. \textbf{(d)}\textbf{ Milling}: All agents revolve around the centroid of the swarm, but they do not form a perfect circle like cyclic pursuit. \textbf{(e)} \textbf{Wall-Following}: Agents slide along the walls and trace the outer edge of the environment.
    }
    \label{fig:discovered-behaviors}
\end{figure*}

\subsection{Robot Simulator}\label{sec:exp-rsrs}
Our HeRo+ robots have a maximum linear velocity of 20~cm/s and a maximum angular velocity of 3~rad/s. However, we noticed that at high-speeds, the robots did not have sufficient time to sense other robots; follow-up tests indicated that the robots had near-perfect sensing at max speeds of 9~cm/s and 1.6~rad/s, which is what we chose to implement in simulation so that our robots could switch velocity commands with sufficient reaction time. In the context of RSRS design, this was a much cheaper way to bridge the reality gap than upgrading the robots with increased sensing frequency or attempting to model the sensing reliability as a function of velocity.

Based on the interactions between robots and the environment we also observed that friction between the robots and the wall was a major aspect that impacted the robot's behavior. Notably, in the default simulator, the collisions are frictionless, which allow robots to slide along walls and other robots with ease. In the real world, however, even though our bump shields allowed for safe collisions, those collisions still involve friction. Therefore, we ran intentional collision tests on our robots and manually approximated the friction coefficient of robot-to-robot and robot-to-wall collisions until our simulator visually matched our observations in the real world. This approach successfully prevents the robots from relying on frictionless interactions to form emergent behaviors. 


\subsection{Representation Learning}\label{sec:exp-repr-learning}
\subsubsection{Training Data} Given our RSRS simulator, we randomly sample unicycle controllers from the constrained space of control velocities measured on the real world robots to create 6000 training videos. To reduce the size of the training data, we render all simulation in greyscale and we also resize the original simulation resolution from 513x426 to 64x64. Each simulation runs for a fixed duration of 600 timesteps ($dt$=0.1). 
To improve processing and training speed, we sub-sample each video to form an input of size (3, 64, 64), where the channel dimension represents 3 greyscale images evenly spaced over the last 300 timesteps of simulation, which we qualitatively found to capture the final converged emergent behavior.

For the random input transformations, we apply a random crop that scales the image in the range $[0.6, 1.0]$ with a 1:1 aspect ratio, a horizontal flip is then applied with probability $p$=0.5. For random rotation, we select a random rotation angle $\theta \in \{0, \frac{\pi}{2}, \pi, \frac{3\pi}{2}\}$ and rotate the axes of the video around the image center.

\subsubsection{Deep Learning} 
We instantiate our representation model, $\phi$, as a pretrained ResNet18~\cite{he2016deep} model with a modified final output of size 128. All but the last layer of the ResNet represent the encoder, and a final 2-layer MLP is used to project the embedding into a space where the loss is applied. The latent embedding is a vector of size 512, which reflects the ResNet's default layer size. As recommended in prior work~\cite{chen2020simple}, we only use the encoder part of the network for downstream evaluation. We train for 100 epochs with a mini-batch size of 1000 videos, following the large-batch recommendations of SimCLR~\cite{chen2020simple}. Each video is passed through the random crop and random rotation transformations to produce a total of 1000 video pairs. For training, we use the NX-Ent loss~\cite{chen2020simple} from Equation~\ref{eq:NXent} paired with the LARS optimizer~\cite{you2017large} with a learning rate of $1.17 = (0.3 \times \text{BatchSize} / 256)$ and weight decay of $1.5\times10^{-7}$. All other hyperparameters follow the original SimCLR implementation~\cite{chen2020simple}.

\subsubsection{Baseline: Hand-Crafted Metrics}
We compare our representation learning to the set of hand-crafted behavior features used by prior work on robots of this same capability model (i.e., differential drive with line-of-sight sensing)~\cite{brown_discovery_2018}. Each metric captures a scalar-valued characteristic of the collective motion of the agents including average speed, angular momentum, radial variance, scatter, and group rotation. When concatenated, the five metrics form a behavior representation $b \in \mathds{R}^5$.

\subsection{Behavior Discovery}\label{sec:exp-behavior-discovery}
Inspired by prior work~\cite{ mattson_leveraging_2023}, we use a tournament-style genetic algorithm to evolve our controllers under the objective function in Equation~\ref{eq:objective_func}. We start with an initial population of 50 controllers randomly sampled from the controller space and run the evolutionary search for 100 generations, each with a population of 50 genomes. At the end of every generation, the resulting behavior videos from each controller are passed through the encoder (or evaluated with the baseline metrics) and saved to the buffer for use in novelty search. Following prior work~\cite{mattson_leveraging_2023}, we compute novelty (Equation~\ref{eq:novelty}) with respect to the 15 nearest-neighbors in the buffer and use a same crossover rate of 0.7 and a mutation rate of 0.15. 
Novelty search results in a buffer of 5000 controllers. The representations of these behaviors are then clustered using k-Medoids with k=10, resulting in 10 behaviors (medoids) from the search that are selected for evaluation in the real world.

\subsection{Real-World Deployment}\label{sec:exp-robot-deploy}
As we did not explicitly filter our behavior space before clustering, it is likely that some behaviors found in behavior discovery will show agents crashing into walls, not moving, or not producing a collective behavior; we refer to these behaviors as \textit{Random} behaviors. We note that after behavior discovery, it is up to the discretion of the human to determine which behaviors they want to try to reproduce on the robots in the real world. For a fair evaluation, we reproduce all non-random behaviors in the real world to assess the success of our RSRS simulator design. 

For each experiment, we record whether or not the behavior was successfully reproduced in the real world and any adjustments to initial conditions or controllers that were required to produce the behavior. It should be noted that our assessment of behavior reproduction is only considered with respect to the swarm's high-level behavior. As there will always be uncertainties in the real world that cannot be modeled in simulation, we do not expect perfect agent-level sim2real alignment and we do not measure how accurately each robot follows its individual simulated trajectory. Rather, our goal is to bring our simulator close enough to reality that swarm-level behavior can be reliably reproduced on real robots.

\section{Results}
Our results average 3 runs of behavior discovery for both the baseline hand-crafted metrics and our self-supervised learned representation. Across both methods, k-Medoids returns 30 non-random controllers for deployment and evaluation in the real world. We show that our automated discovery can detect emergent behaviors of Cyclic Pursuit (Cyc.), Aggregation (Agg.) and Dispersal (Disp.) that can be deployed directly into the real world (Figure~\ref{fig:discovered-behaviors} a-c). We first analyze the performance of our representation learning, then discuss the behaviors that were discovered and successfully deployed onto real-world robots. We also highlight results from an additional study which supports the inclusion of RSRS design in our approach, where we show two other behaviors, Milling and Wall-Following (see Figure~\ref{fig:discovered-behaviors} d-e), that are also discovered by a naive behavior search approach that does not use RSRS. While interesting, these emergent behaviors cannot be deployed into the real world, whereas all the emergent behaviors discovered via our approach are directly deployable on our real robots.

\subsection{Representation Alignment}
\begin{figure}[t!]
    \centering
    \begin{subfigure}[t]{0.45\linewidth}
        \centering
        \includegraphics[width=0.9\linewidth]{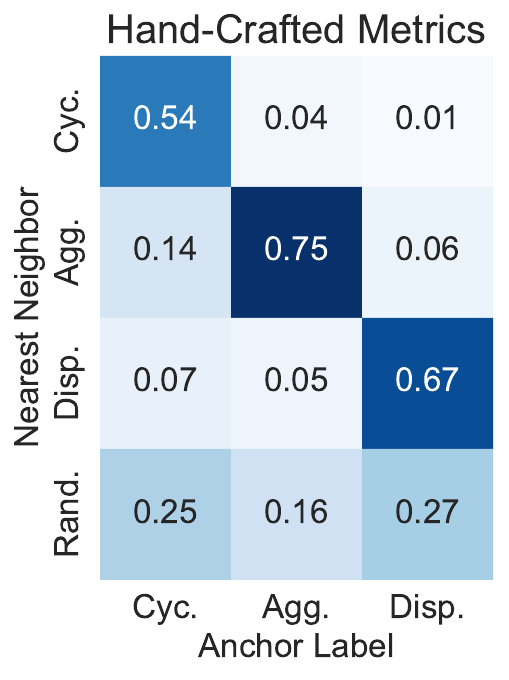}
    \end{subfigure}%
    \hfill
    \begin{subfigure}[t]{0.45\linewidth}
        \centering
        \includegraphics[width=0.9\linewidth]{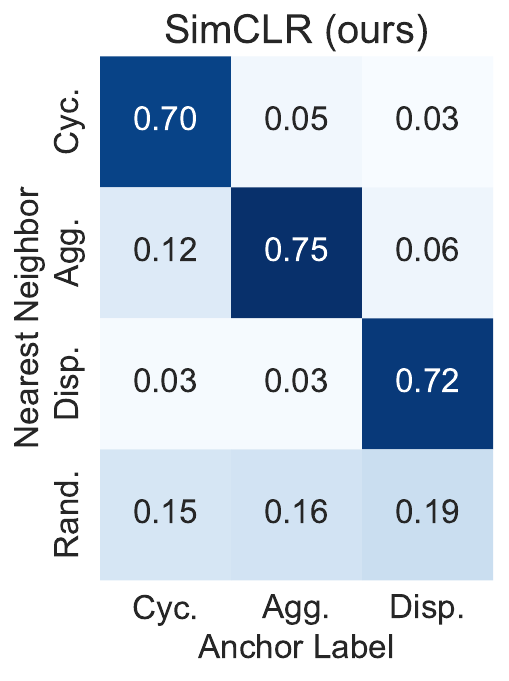}
    \end{subfigure}
    \caption{
    Representation Confusion Matrices
    \normalfont for \textbf{(left)} the baseline hand-crafted representations and \textbf{(right)} the self-supervised representations on a held-out set of 500 labeled test videos. Classes along the horizontal axis indicate the labeled class of an anchor behavior and the values along the vertical axis display the frequency with which a behavior from the anchor's class (on-diagonal) was embedded closer to the anchor when compared with a behavior from one of the other classes (off-diagonal). Larger diagonal values indicate stronger within-class correlation in the embedded representation space. 
    }
    \label{fig:res-class-confusion}
\end{figure}

\begin{figure*}[t!]
    \centering
    \begin{subfigure}[t]{0.43\linewidth}
        \centering
        \includegraphics[width=\linewidth]{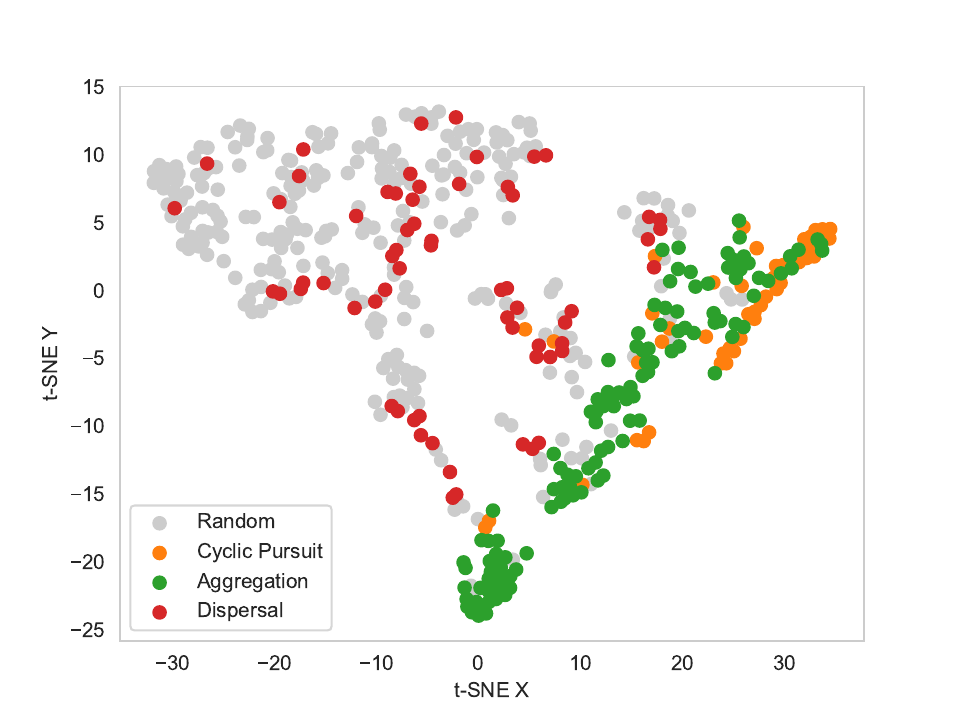}
        \caption{Hand-Crafted Behavior Metrics~\cite{brown_discovery_2018}}
    \end{subfigure}%
    ~ 
    \begin{subfigure}[t]{0.43\linewidth}
        \centering
        \includegraphics[width=\linewidth]{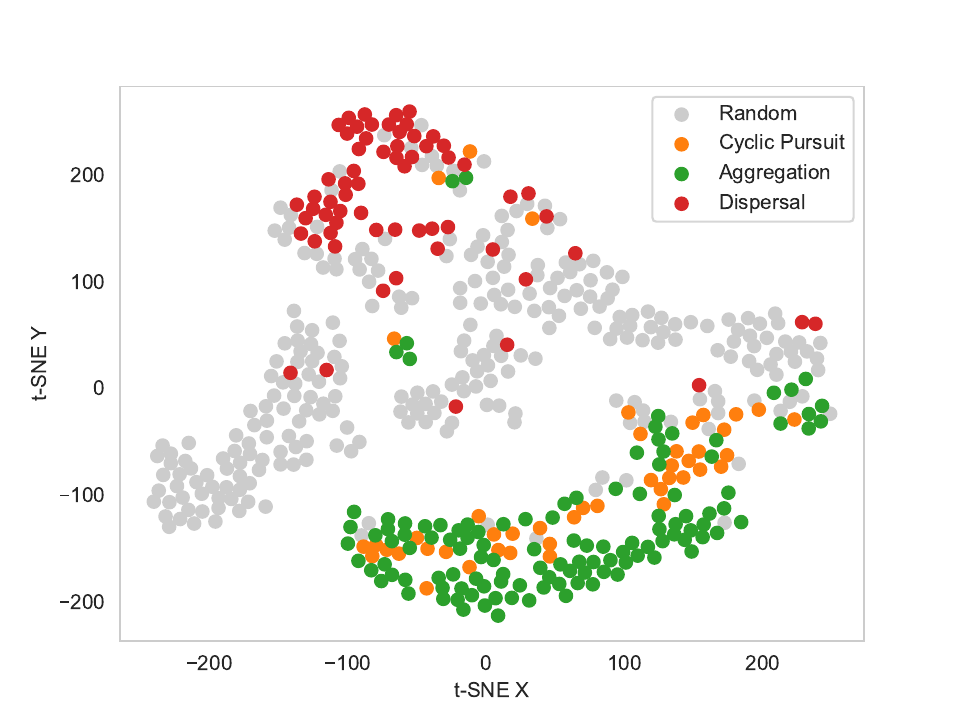}
        \caption{Self-Supervised Representation Learning (ours)}
    \end{subfigure}
    \caption{
    t-SNE Embeddings
    \normalfont for the \textbf{(a)} baseline hand-crafted representations and the \textbf{(b)} self-supervised representations on a held-out set of 500 labeled test videos. Qualitatively, the hand-crafted baseline is able to densely associate cyclic pursuit behaviors, but fails to differentiate dispersal from random behaviors. In our approach, behavior features have more variance, but are more closely associated with behaviors from the same class, notably dispersal, which is embedded more distinctly from random behaviors than in the baseline case.
    }
    \label{fig:res-tSNE}
\end{figure*}

Although we did not require any labeled data for training, we evaluate how well our encoder performs using a set of 500 labeled testing videos. We first compute a set of representation-specific confusion matrices (Figure~\ref{fig:res-class-confusion}), which indicate how often different examples of the same emergent behavior were embedded closer together than a different emergent behavior. 
To evaluate the quality of our learned latent representation, we utilize a triplet, $\left<a, p, n\right>$, consisting of an anchor ($a$), positive ($p$), and negative ($n$) example, where $a$ and $p$ have the same label and $n$ is a different label than both $a$ and $p$. Under a correct learned representation, we would expect the embedded representations $a$ and $p$ to be more similar in the embedding space than the embedding of $n$ when compared to $a$ or $p$. We adapt the triplet distance formation to create a representation confusion matrix, where we evaluate all valid triplets in the labeled testing data defined by 
    $\left\{\left<a, p, n\right> \subset X_{\text{test}} \mid a_{\text{label}} = p_{\text{label}},  a_{\text{label}} \neq n_{\text{label}}\right\}$,
and then test to see if $dist(a, p) < dist(a, n)$. We conduct this test for both the baseline and self-supervised representation methods. We evaluate the static hand-crafted metrics with a single evaluation and the learned representation as the averages of 3 SimCLR training runs. 

Figure~\ref{fig:res-class-confusion} shows our findings for emergent behavior representation alignment, where we see that both methods display a clear correlation within-behavior as shown by the high values on the diagonal. We find that, although the baseline metrics were specifically crafted to reflect swarm behavior characteristics, the self-supervised capabilities of SimCLR are able to perform with slightly better or equal accuracy than the baseline for all non-random behaviors (+16\% cyclic pursuit, +0\% aggregation, +5\% dispersal), indicating that \textit{the learned model can sufficiently capture behavioral semantics using only self-supervised representation learning}, and that it outperforms the hand-crafted approach.  


We also evaluate the representation of the labeled data from a qualitative perspective by visualizing the embeddings in 2D space using t-SNE dimensionality reduction~\cite{van2008visualizing}. The hand-crafted metrics (Figure~\ref{fig:res-tSNE}a) show a clear ability to distinguish random behaviors from cyclic pursuit and aggregation. However, the hand-crafted representation poorly differentiates between dispersal and random controllers, which supports the findings of our quantitative data (Figure~\ref{fig:res-class-confusion}a) that show that hand-crafted dispersal representations are confused for random representations 27\% of the time. Notably, in Figure~\ref{fig:res-tSNE}b, the learned embedding does still correctly differentiate aggregation and cyclic-pursuit from random, but performs with 8\% better accuracy when separating dispersal from random.
Importantly, in both representation confusion and t-SNE evaluations, our self-supervised model is being evaluated on data that was not seen during training, indicating a strong ability to generalize.

\begin{table}[]
\small
\centering
\caption{
Discovered Behavior Frequency 
\normalfont for novelty search with k=10 behaviors returned in each trial. Results are averaged across 3 runs and displayed alongside the standard error.}
\begin{tabular}{lccc}
                                     & \multicolumn{3}{c}{\textbf{Behavior Representation}}                    \\ \cline{2-4} 
\multicolumn{1}{l|}{\textbf{\begin{tabular}[c]{@{}l@{}}Discovered \\ Behaviors\end{tabular}}} &
  \begin{tabular}[c]{@{}c@{}}Hand\\ Crafted~\cite{brown_discovery_2018}\end{tabular} &
  \begin{tabular}[c]{@{}c@{}} Triplet\\ Learning~\cite{mattson_leveraging_2023}\end{tabular} &
  \multicolumn{1}{c|}{\begin{tabular}[c]{@{}c@{}}Self-Supervised\\ (ours)\end{tabular}} \\ \hline
\multicolumn{1}{|l|}{Aggregation}    & 4.33 $\pm$ 0.66 & 3.66 $\pm$ 0.66 & \multicolumn{1}{c|}{2.0 $\pm$ 0.0}  \\
\multicolumn{1}{|l|}{Cyclic Pursuit} & 0.66 $\pm$ 0.33 & 2.0 $\pm$ 0.0   & \multicolumn{1}{c|}{1.0 $\pm$ 0.57} \\
\multicolumn{1}{|l|}{Dispersal}      & 0.0 $\pm$ 0.0   & 0.0 $\pm$ 0.0   & \multicolumn{1}{c|}{2.0 $\pm$ 0.57} \\
\multicolumn{1}{|l|}{Random}         & 5.0 $\pm$ 1.0   & 4.33 $\pm$ 0.66 & \multicolumn{1}{c|}{5.0 $\pm$ 1.0}  \\ \hline
\multicolumn{1}{|l|}{\begin{tabular}[c]{@{}l@{}}Total Unique\\ (excl. Random)\end{tabular}} &
  1.6 $\pm$ 0.33 &
  2.0 $\pm$ 0.0 &
  \multicolumn{1}{c|}{\textbf{2.6 $\pm$ 0.33}} \\ \hline
\end{tabular}
\label{tab:res-behvior-freq}
\end{table}

\begin{table}[]
\caption{Real2Sim2Real (RSRS) Experiments
\normalfont where our self-supervised method is compared with and without the RSRS~\cite{vega_simulate_2023} simulator improvements in 3 trials of behavior discovery. Although the default simulator discovered additional behaviors, they could not be produced in the real world. Legend: (\cmark) Behavior discovered and one-shot deployed on real robots, (\halfcmark) behavior discovered and deployed with 2-3 attempts, (\xmark) behavior discovered, but not successfully deployed, (--) behavior not discovered. Multiple entries per cell indicate that the behavior was returned from behavior discovery more than once.}
\centering
\small
\begin{tabular}{|l|ccc|ccc|}
\hline
 &
  \multicolumn{3}{c|}{\begin{tabular}[c]{@{}c@{}}\textbf{No RSRS} \\ (Sim. Default)\end{tabular}} &
  \multicolumn{3}{c|}{\begin{tabular}[c]{@{}c@{}}\textbf{RSRS}\end{tabular}} \\ \cline{2-7} 
\begin{tabular}[c]{@{}l@{}}\textbf{Behavior}\end{tabular} &
  \multicolumn{1}{c|}{\begin{tabular}[c]{@{}c@{}}Trial\\ 1\end{tabular}} &
  \multicolumn{1}{c|}{\begin{tabular}[c]{@{}c@{}}Trial\\ 2\end{tabular}} &
  \begin{tabular}[c]{@{}c@{}}Trial\\ 3\end{tabular} &
  \multicolumn{1}{c|}{\begin{tabular}[c]{@{}c@{}}Trial\\ 1\end{tabular}} &
  \multicolumn{1}{c|}{\begin{tabular}[c]{@{}c@{}}Trial\\ 2\end{tabular}} &
  \begin{tabular}[c]{@{}c@{}}Trial\\ 3\end{tabular} \\ \hline
Aggregation &
  \multicolumn{1}{c|}{\halfcmark \xmark} &
  \multicolumn{1}{c|}{\nmark} &
  \xmark \xmark \xmark &
  \multicolumn{1}{c|}{\cmark \cmark} &
  \multicolumn{1}{c|}{\cmark  \halfcmark} &
  \cmark \cmark \\
Cyclic Pursuit &
  \multicolumn{1}{c|}{\xmark} &
  \multicolumn{1}{c|}{\cmark} &
  \cmark \xmark &
  \multicolumn{1}{c|}{\cmark} &
  \multicolumn{1}{c|}{\cmark  \halfcmark} &
  \nmark \\
Dispersal &
  \multicolumn{1}{c|}{\nmark} &
  \multicolumn{1}{c|}{\cmark \cmark} &
  \nmark &
  \multicolumn{1}{c|}{\cmark} &
  \multicolumn{1}{c|}{\cmark \cmark \halfcmark} &
  \cmark \cmark \\
Milling &
  \multicolumn{1}{c|}{\nmark} &
  \multicolumn{1}{c|}{\xmark} &
  \xmark \xmark &
  \multicolumn{1}{c|}{\nmark} &
  \multicolumn{1}{c|}{\nmark} &
  \nmark \\
Wall Following &
  \multicolumn{1}{c|}{\xmark} &
  \multicolumn{1}{c|}{\xmark \xmark} &
  \nmark &
  \multicolumn{1}{c|}{\nmark} &
  \multicolumn{1}{c|}{\nmark} &
  \nmark \\ \hline
\end{tabular}
\label{tab:res-rsrs-ablation}
\end{table}


\subsection{Discovery and Deployment}
We run behavior discovery 3 times for both the baseline metrics and our learned representation. Each time, a set of 10 controllers is output and categorized by behavior. The frequency of returned behaviors is shown in Table~\ref{tab:res-behvior-freq}. We find on average that the number of behaviors extracted from each method appears to correlate very closely with how the behaviors were distributed in the t-SNE visualization, with dispersal never being discovered in the baseline method, as one might hypothesize based on Figure~\ref{fig:res-tSNE}a. Though the other behaviors vary consistently, it is also worth noting that both methods return the same number of random behaviors on average. 

From the 60 controllers examined in simulation, exactly 30 (15 from each method) were identified as non-random emergent behaviors.  We directly deployed the same controllers found in simulation on our HeRo+ robots in the real world. From the 30 non-random controllers discovered in the RSRS simulator, we were able to one-shot reproduce \textbf{70\% (20)} of them in the real world and \textbf{90\% (27)} were successfully reproduced within 3 attempts, without any controller or initialization adjustment. 
Our methods show the potential for sim2real transfer in tandem with behavior discovery using the RSRS method. We validate this further in a thorough ablation study where we examine the importance of each improvement to our simulator by deploying behaviors discovered under simulators with incomplete RSRS measurements.

\subsection{Importance of RSRS in Swarm Deployment}\label{sec:res-ablation}
To demonstrate the importance of RSRS in our approach, we ran 3 additional trials of our self-supervised method on the original simulator, with almost no RSRS improvements. To ensure a fair comparison that has the potential to produce deployable behaviors, we implement the physical dimensions of the original HeRo robots and upper-bound the controller with the robot's maximum forward and angular velocities, but we do not include any of our measurements for friction, reasonable speeds for sensing, or the augmented geometry from the bump shield. Our experiments (Table~\ref{tab:res-rsrs-ablation}) show that a total of 18 non-random controllers were discovered, including two behaviors that were not discovered in our RSRS behavior discovery: wall following and milling. Of the 18 controllers returned from this ablation experiment, only 22\% (4) of the behaviors could be one-shot reproduced on the robots and 27\% (5) were successfully reproduced within 3 attempts. Compared to the default simulator, \textit{the inclusion of RSRS improves the one-shot success rate by 48\% and the three-shot success rate by 63\%.} 

Notably, neither wall following nor milling were reproduced successfully, indicating that the RSRS behavior discovery did not erroneously miss these behaviors---rather, these behaviors are artifacts of an imperfect simulator and are not achievable with the real capabilities of our HeRo+ robots. For wall following, real-world friction prevents agents from sliding along the walls of the environment. Milling could not be achieved in the real world because it leads to many head-on collisions between robots, which cannot easily slip past one another as they can in the unrefined simulator.   



\section{Conclusion and Future Work}
We present Real2Sim2Real Behavior Discovery via Self-Supervised
Representation Learning and show the successful real-world deployment of emergent behaviors discovered in simulation. We also demonstrate that purely self-supervised learned behavioral representations can be used in place of burdensome hand-crafted metrics and outperform the hand-crafted metrics in terms of emergent behavior representation and discovery. 
In the future, we are excited to apply our approach to novel swarm capability models, including limited communication capabilities. We are also excited to study the effect of the environment ($\mathcal{E}$) on the possible emergent swarm behaviors. 
Finally, we believe that behavior discovery has the potential to enable swarm roboticists to discover novel ways of using robots to accomplish interesting and useful tasks, enabling the confident deployment of automatically discovered behaviors into real-world scenarios. Future work should study specific tasks where novelty and discovery can improve how we solve problems with scalable swarms of low-cost, limited-capability robots.



\begin{acks}
This work was conducted in the Aligned, Robust, and Interactive Autonomy (ARIA) Lab at the University of Utah. ARIA Lab research is supported in part by the NSF (IIS-2310759, IIS2416761), the NIH (R21EB035378), ARPA-H, and the ARL STRONG program.
\end{acks}



\bibliographystyle{ACM-Reference-Format} 
\bibliography{main}


\end{document}